\documentclass[letterpaper, 10 pt, conference]{ieeeconf}  %

\IEEEoverridecommandlockouts                              %

\usepackage{graphics} %
\usepackage{epsfig} %
\usepackage[bookmarks=true]{hyperref}
\usepackage[natbib=true,
    style=ieee,
    citestyle=numeric-comp,
    backend=biber,
    maxbibnames=10,
    maxcitenames=99,
    doi=false,
    url=false,
    giveninits=true]{biblatex}
\renewcommand*{\bibfont}{\footnotesize}

\usepackage{comment}
\usepackage{amssymb}
\usepackage{amsmath}
\usepackage{multirow}
\usepackage{multicol}
\usepackage{balance}
\usepackage{siunitx}

\usepackage[ruled,vlined,linesnumbered]{algorithm2e}
\SetInd{0.1em}{0.4em}
\SetAlFnt{\footnotesize\sf}
\SetKwRepeat{Do}{do}{while}
\SetKw{KwStep}{step}
\SetKwInput{KwData}{Input}
\SetKwInput{KwResult}{Result}
\SetKwBlock{RepeatForever}{repeat}{}
\SetKw{Continue}{continue}
\SetKwComment{Comment}{$\triangleright$\ }{}
\SetCommentSty{itshape}

\usepackage[capitalise]{cleveref}
\crefname{equation}{}{} %

\usepackage{array}

\usepackage{tikz}
\usetikzlibrary{fit,calc}
\newcommand{\Fd}{\mathbf{F}_d}

\newcommand{\mud}{\boldsymbol \mu_{i_d}}

\newcommand{\normmu}{\|\boldsymbol \mu_{i_d}\|}

\newcommand{\qio}{\mathbf{q}_{i_r}}
\newcommand{\dotqio}{\dot{\mathbf{{q}}}_{i_r}}
\newcommand{\wio}{\boldsymbol{\omega}_{i_r}}
\newcommand{\dotwio}{\dot{\boldsymbol{\omega}}_{i_r}}

\newcommand{\qidot}{\dot{\mathbf{q}}_i}
\newcommand{\qi}{\mathbf{q}_i}
\newcommand{\qihat}{\hat{\mathbf{q}}_i}
\newcommand{\wi}{\boldsymbol{\omega}_i}
\newcommand{\wdot}{\dot{\boldsymbol{\Omega}}}
\newcommand{\wuav}{\boldsymbol{\Omega}}

\newcommand{\qibase}{\mathbf{q}_{i_{\text{base}}}}

\newcommand{\qoneo}{\mathbf{q}_{1_r}}
\newcommand{\qno}{\mathbf{q}_{n_r}}

\newcommand{\muo}{\boldsymbol \mu_{i_r}}

\newcommand{\po}{\mathbf{p}_0}

\newcommand{\vo}{\dot{{\mathbf{p}}}_0}

\newcommand{\por}{\mathbf{p}_{0_r}}
\newcommand{\p}{\mathbf{p}}
\newcommand{\podr}{\dot{\mathbf{p}}_{0_r}}
\newcommand{\poddr}{\ddot{\mathbf{p}}_{0_r}}
\newcommand{\podd}{\ddot{\mathbf{p}}_{0}}

\newcommand{\ez}{\mathbf{e_3}}
\newcommand{\R}{\mathbf{R}}

\newcommand{\C}{\mathcal{C}}
\newcommand{\W}{\mathcal{W}}
\newcommand{\sS}{\mathcal{S}}

\newcommand{\X}{\boldsymbol{X}}
\newcommand{\U}{\boldsymbol{U}}
\newcommand{\x}{\boldsymbol{x}}
\newcommand{\uinp}{\boldsymbol{u}}

\newcommand{\xconfg}{\boldsymbol{{x}}_{geom}}

\newcommand{\az}{ \alpha}
\newcommand{\el}{ \gamma}

\title{\LARGE \bf
Kinodynamic Motion Planning for a Team of Multirotors Transporting a Cable-Suspended Payload in Cluttered Environments
}

\author{Khaled Wahba$^1$, Joaquim Ortiz-Haro$^{1,2}$, Marc Toussaint$^1$, and Wolfgang Hönig$^1$%
\thanks{$^1$ Faculty of Electrical Engineering and Computer Science,
        Technical University of Berlin, Berlin, Germany
        {\tt\footnotesize k.wahba@tu-berlin.de}.}%
\thanks{$^2$ Machines in Motion Laboratory, New York University, USA}%
\thanks{Code: \url{https://github.com/IMRCLab/coltrans-planning}}%
\thanks{Website: \url{https://imrclab.github.io/coltrans-planning/}}
\thanks{The research was funded by the Deutsche Forschungsgemeinschaft (DFG,
German Research Foundation) - 448549715.}
}

\begin{document}

\maketitle
\thispagestyle{empty}
\pagestyle{empty}

\begin{abstract}
We propose a motion planner for cable-driven payload transportation using multiple unmanned aerial vehicles (UAVs) in an environment cluttered with obstacles. Our planner is kinodynamic, i.e., it considers the full dynamics model of the transporting system including actuation constraints. Due to the high dimensionality of the planning problem, we use a hierarchical approach where we first solve for the geometric motion using a sampling-based method with a novel sampler, followed by constrained trajectory optimization that considers the full dynamics of the system. Both planning stages consider inter-robot and robot/obstacle collisions. We demonstrate in a software-in-the-loop simulation and real flight experiments that there is a significant benefit in kinodynamic motion planning for such payload transport systems with respect to payload tracking error and energy consumption compared to the standard methods of planning for the payload alone. Notably, we observe a significantly higher success rate in scenarios where the team formation changes are needed to move through tight spaces.
\end{abstract}

\section{Introduction}

Uncrewed aerial vehicles (UAVs) are ideal for tasks that involve accessing remote locations, which makes them valuable collaborators in a variety of scenarios.
Cable-driven payload transportation using multiple UAVs is well suited for
collaborative assistance in construction sites such as carrying tools \cite{gabellieri2018study} or transporting materials.

The field of control methods for payload transport systems has witnessed significant advancements. In particular, control algorithms \cite{lee2017geometric} 
have been devised to solve the transport problem with stability guarantees.
However, they  neglect inter-robot and robot/obstacle collisions.
Conversely, alternative methods employ a nonlinear optimization framework to account for inter-robot collisions, which require intricate online and on-board computations.

A common limitation of the current methods for payload transport systems is that they assume a provided feasible reference trajectory that can be tracked. 
However, generating such reference trajectories for nonlinear, high-dimensional systems in cluttered environments is a recurring challenge and still an open problem. 
Traditionally, reference trajectories have been computed using either linear interpolation, planning for simplified dynamical models, or planning only for the payload.
However, when the controller attempts to track these dynamically unfeasible trajectories in cluttered environments, it is likely to fail from either motor saturation or collisions caused by high tracking errors.
These effects are more notable if agile maneuvers are desired, or robots with low thrust-to-weight ratio are employed.
\begin{figure}[t]
    \centering
    \includegraphics[width=\linewidth]{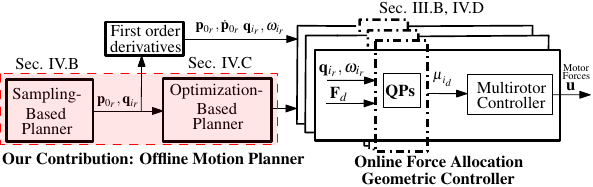}
    \caption{Highlighted in the red box is our full kinodynamic motion planning algorithm. The geometric output of a sampling-based motion planner is used to initialize an optimizer, which generates the full feasible reference trajectory of the payload $\por$ and the cable states $\qio$. One can also use our sampling-based motion planner and compute the first order derivatives of the geometric states $(\podr,\wio)$ to provide a reference trajectory. Each reference trajectory is then tracked by our controller \cite{wahba2023efficient}.
    }
    \label{fig:high-level-arch}
    \vspace{-5pt}
\end{figure}
To the best of our knowledge, there is no kinodynamic motion planner for cable-suspended payload transport.

We show that such a planner has significant advantages over traditional motion planning methods, as it is possible to construct feasible trajectories for the entire system's state, including not only the payload but also the UAVs and cable states. These trajectories account for inter-robot, robot/obstacle, cable/obstacle collisions, and the actuation limits of the motors.
Overall, tracking such trajectories with existing controllers can lead to more reliable operation (success rate), higher predictability (lower tracking error), and lower energy consumption as flight time can be reduced.

In this paper, we extend our previous work \cite{wahba2023efficient} by proposing an offline hierarchical kinodynamic motion planner (see \cref{fig:high-level-arch}) to generate feasible reference trajectories to transport a point mass with multiple aerial multirotors in environments cluttered with obstacles.
We use an enhanced version of our prior geometric sampling-based motion planner \cite{wahba2023payloadplanning} as an initial guess for a nonlinear optimizer. The optimizer generates feasible energy-efficient reference trajectories that take 
collisions
into account. 
We evaluate the planner by tracking reference trajectories using our highly-efficient controller \cite{wahba2023efficient} and compare it using a realistic software-in-the-loop (SITL) simulation and several real flights with two baselines: the geometric cable-payload planner and the payload-only planning baseline. We consider three different environments, vary the number of robots and report key metrics such as energy efficiency, tracking accuracy, and success rate.

\section{Related Work}

\textbf{Control algorithms} of the payload transport system include centralized controllers, employing a cascading reactive approach with stability guarantees \cite{lee2017geometric,lee2013geometric, six2017kinematics} and decentralized controllers \cite{tagliabue2019robust,tognon2018aerial}. There are still practical obstacles to overcome, including the requirement to measure noisy payload accelerations and considering inter-robot collisions.

Optimization-based controllers can directly include some constraints. One approach involves using iterative gradient-based solvers, but they are susceptible to local minima \cite{jimenez2022precise, petitti2020inertial}. 
Nonlinear model predictive control (NMPC) offers an alternative for payload control and collision avoidance \cite{li2023nonlinear, sun2023nonlinear}. However, its high computational costs and limited scalability with multiple robots make it unsuitable for resource-constrained microcontrollers. Moreover, these methods do not directly integrate the dynamic model of the multirotors or the actuation limits of their motors into their online optimization formulation.
They rely only on cable tension constraints and state bounds, which are only suitable for slow-varying trajectories.
Our previous work \cite{wahba2023efficient, wahba2023payloadplanning} combines advantages of prior work by leveraging QPs that can handle inter-robot, robot/obstacle and cable/cable collision constraints. We proposed a QP-force allocation geometric controller that is executed on compute-constrained multirotors in realtime efficiently. 

Other methods employ offline \textbf{motion planners} \cite{manubens2013motion, de2019flexible, zhang2023if} for inter-robot and robot/obstacle collision avoidance. These planners use sampling-based or control-based approaches, but often overlook multirotor actuation limits, state bounds, and payload distribution, potentially suggesting impractical configurations for the controller.
From a broader robotics perspective, kinodynamic motion planning can rely on search or sampling~\cite{li2016asymptotically,pivtoraikoKinodynamicMotionPlanning2011,webb2012kinodynamic, 
honig2022db}. Yet, these methods scale exponentially with the dimensionality of the state space, and thus fail when planning for a cable-suspended payload system.

In contrast, constrained trajectory \textbf{optimization} methods \cite{Crocoddyl, malyutaConvexOptimizationTrajectory2022a, TrajOpt,howell2019altro,pardo2016evaluating, geng2022load}
are more suitable for planning in high dimensional state spaces, with polynomial complexity on the state size. 
Kinodynamic motion planning using nonlinear optimization has shown great success in different robotics fields, from sequential manipulation planning \cite{toussaint2018differentiable}
to legged locomotion \cite{ponton2018time, carpentier2018multicontact,winkler2018gait}
and flying robots \cite{foehn2017fast,geisert2016trajectory}.
However, optimization methods require a good initial guess of the solution trajectory, as they can only optimize the trajectory locally. 
In our method, we combine a novel geometric sampling-based motion planner for cable-suspended payload systems with multiple multirotors with subsequent nonlinear trajectory optimization for full kinodynamic planning, resulting in a state-of-the-art integrated system.

\section{Background}
This section provides necessary background for the dynamic model and the used control design, see~\cite{wahba2023efficient} for details. 

\subsection{System Description} \label{sec:dynamics}
The \textbf{dynamic model} of the cable-suspended payload system with multiple multirotors can be described through Lagrangian mechanics \cite{lee2017geometric,masone2016cooperative,pereira2017control,tuci2018cooperative}.
Consider a team of $n$ multirotors transporting a cable-suspended payload. 
The payload is a point mass with mass $m_0$ and the cables are massless rigid rods each with length $l_i$, where $i \in \{1,\hdots, n\}$. 
Each UAV is modeled as a floating rigid body with mass $m_i$ and diagonal moment of inertia $\mathbf{J}_i$.

The payload state is defined by the position and velocity vectors $\mathbf{p}_0\in \mathbb{R}^3$ and  $\vo \in \mathbb{R}^3$. While the cable states are composed of the cable unit vector $\qi \in \mathbb{S}^2$ directed from the multirotor towards the payload, where $\mathbb{S}^2 = \{\mathbf{q} \in \mathbb{R}^3 \big| \|\mathbf{q}\| = 1\}$, and the cable angular velocity $\boldsymbol{\omega}_i \in \mathbb{R}^3$. 
Moreover, the multirotor position and velocity state vectors are $\p_i\in \mathbb{R}^3$ and $ \dot{\p}_i \in \mathbb{R}^3$ with respect to the global frame of reference. As the cables are modeled rigidly, the position of each multirotor can be computed as 
\begin{equation}
    \label{eq:uavpos}
    \p_i = \po - l_i\qi.
\end{equation}
The attitude states of the \emph{i}-th multirotor is comprised of the rotation matrix $\R_i \in SO(3)$ and body angular velocity $\wuav_i \in \mathbb{R}^3$. 
In summary, the configuration manifold $\C$ of the presented system is defined by $\mathbb{R}^3  \times (\mathbb{S}^2 \times SO(3))^n$, where the full system state is defined by
\begin{equation}
    \label{eq:statespace}
    \x = (\po, \vo, \mathbf{q}_1, \boldsymbol{\omega}_1, \mathbf{R}_1, \mathbf{\Omega}_1, \ldots, \mathbf{q}_n, \boldsymbol{\omega}_n, \mathbf{R}_n, \mathbf{\Omega}_n)^T.
\end{equation} 
The output wrench of the \emph{i}-th robot is defined by the collective thrust and the torques $\boldsymbol{\eta}_i = (f_{ci},\mathbf{M}_i)^T$, where $\mathbf{M}_i =(\tau_{xi},\tau_{yi},\tau_{zi})^T$.  This wrench vector is linearly related to the control input motor forces $\uinp^i$ of the multirotor by $\boldsymbol{\eta}_i = \boldsymbol{B}_{0}\uinp^i$, where $\boldsymbol{B}_{0}$ is the actuation matrix and the motor force command is $\uinp^i = (f_{i_1}, f_{i_2},f_{i_3},f_{i_4})^T$.
The control input vector of the full system is defined as 
\begin{equation}
    \label{eq:motorforces}
    \uinp= (\uinp^1, \hdots, \uinp^n)
\end{equation} 
and the system kinematics and dynamics are
\begin{align}
    \label{eq:dynamics}
    &\qidot = \wi \times \qi,\\
    &\mathbf{M_{t}}(\podd +g\ez)  = \sum_{i=1}^n (f_{ci}\R_i\ez - m_il_i\|\wi\|^2\qi), \nonumber\\
    &m_il_i\dot{\boldsymbol{\omega}}_i = m_i\qihat(\podd +g\ez) - f_{ci}\qihat\R_i\ez, \nonumber\\
    &\dot{\R}_i = \R_i \hat{\boldsymbol{\Omega}}_i,
    \quad \mathbf{J}_i\wdot_i = \mathbf{J}_i\wuav_i \times \wuav_i + \mathbf{M}_i, \nonumber
\end{align}
where $\podd$ is the acceleration of the payload, $g$ is the gravitational acceleration constant, $\ez = (0,0,1)^T$ and  $\mathbf{M}_{t} = (m_0 + \sum_{i=1}^n m_i)\mathbf{I}_{3\times3}$. $(\hat{\cdot})$ denotes the skew-symmetric mapping $\mathbb{R}^3 \rightarrow \mathfrak{s}\mathfrak{o} (3)$. 
\subsection{Controller Overview}

Our control design (see \cref{fig:high-level-arch}) presents an efficient optimization-based cable force allocation of a geometric controller \cite{lee2017geometric} for cable-suspended payload transportation that is aware of neighboring robots to avoid collisions.
Consider the desired control forces that track the payload reference trajectory as $\Fd$.
This method reformulates the cable forces allocation optimization problem as two consecutive quadratic programs (QPs). 
The QPs solve for the desired cable forces $\mud$, taking into account the inter-robot collisions, and track $\Fd$. 
Thus, $\mud$ is tracked by the \emph{i}-th multirotor with a low-level controller \cite{lee2013geometric}.

\section{Approach}
\subsection{Problem Statement}
Consider the system described in \cref{sec:dynamics}. The state space vector is defined by \eqref{eq:statespace}. 
Let $\X = \langle \x_0, \x_1, \hdots, \x_T \rangle$ be a sequence of states sampled at time $0, \Delta t, \hdots, T\Delta t$ and $\U = \langle \uinp_0, \uinp_1, \hdots, \uinp_{T-1} \rangle$ be a sequence of controls applied to the system for times $[0, \Delta t), [\Delta t, 2\Delta t) \hdots, [(T-1)\Delta t, T\Delta t)$, where $\Delta t$ is a small timestep and the controls are constant during this timestep. 
We denote the start state as $\x_s$, the goal state as $\x_g$ and the collision-free configuration space as $\C_{\text{free}} \subset \C$, which accounts for collisions between the robots, payload, and the cables as well as collisions against the environment.
Then our goal is to transport the payload from a start to a goal state in the optimal time $T$, which can be framed as the following optimization problem
\begin{align}
    &\min_{\X, \U, T} \hspace{0.2cm} J(\X, \U, T), \label{eq:general-optimization-problem}\\
    &\text{\noindent s.t.}\begin{cases}
     \x_{k+1} = \text{step}(\x_k, \uinp_k) \quad \forall k\in\{0, \ldots, T-1\}, \nonumber \\
     \uinp_k \in \mathcal{U}  \quad \forall k\in\{0, \ldots, T-1\}, \nonumber \\
     \x_0 = \x_s, \hspace{0.2cm} \x_T = \x_f, \nonumber \\
      \x_k \in \C_{\text{free}} \subset \mathcal{C}  \quad \forall k\in\{0, \ldots, T\}, \\
    \end{cases}
\end{align}
where the cost function can be set to minimize $T$ and other task objectives (e.g., energy). 
The function 
$\text{step}(\x_k, \uinp_k)$ is the time-discretized version of the dynamic model of the system and the second constraint limits the control input within a feasible control space $\mathcal{U}$ (e.g., actuator limits). 
The third set of constraints ensures that the motion connects the start and the goal states. 
The last constraint ensures a collision-free path for the full state while avoiding any cable entanglements. 

\subsection{Geometric Motion Planning (Offline)}
\label{sec:approach:geometric}
Given a start state $\x_s$ and a goal state $\x_g$ in an environment 
with obstacles, we propose to use a sampling-based motion planner to plan a collision-free geometric path $\X_\text{geom}$ for the following state vector
\begin{equation}
\x_\text{geom} = (\por, \qoneo, \hdots, \qno)^T,
\end{equation}
where $\x_\text{geom} \in \mathbb{R}^{3} \times \mathbb{R}^{3n}$ and subscript $r$ is the reference trajectory. 
Consequently, $\X_\text{geom}$ is interpolated and the first order derivatives are computed by numerical differentiation to define the full reference trajectory that can be tracked by a controller or used as an initial guess for an optimizer.  

\subsubsection{State Space Representation}
We propose to reduce the state space size directly by using a local parametrization for the unit vector $\qi \in \mathbb{R}^3$ with azimuth $\az_i$ and elevation $\el_i$ angles, such that
\begin{equation}
    \label{eq:azeltoqi}
    \qi = -\big(\cos(\az_i)\cos(\el_i), \hspace{0.1cm} \sin(\az_i)\sin(\el_i), \hspace{0.1cm} \sin(\el_i)\big)^T,
\end{equation}
where $\az_i \in [0, 2\pi)$ and $\el_i \in [0, \pi/2)$. 
As the payload is always beneath the robots, there are no singularities in this representation.
Thus, the reduced state vector can be represented by
\begin{equation}
\xconfg = (\por, \az_{1_r}, \el_{1_r}, \hdots, \az_{n_r}, \el_{n_r})^T \in \mathbb{R}^{3} \times \mathbb{R}^{2n}.
\end{equation}
\subsubsection{Cost Function}   
Given two consecutive states as $\xconfg(k)$ and $\xconfg(k+1)$, we use a cost function that minimizes the integral of the energy between the two states. First, let us define the $\qibase = (0,0,-1)^T$ as the default cable direction to carry the payload statically. 
Consider the required force magnitude to carry a unit payload mass with respect to the static case (i.e., all cables point towards $\qibase$) as
\begin{equation}
    F(k) = -\frac{1}{n}\sum_{i=1}^n \frac{1}{\qibase^T\qi(k)},
\end{equation}
where $n$ is the number of cables. We assume a trapezoidal energy profile between two consecutive states. Thus, the cost function is 
\begin{equation}
    c = \frac{F(k)+F(k+1)}{2}(\beta\hspace{0.2cm}\p_{0}^{\delta} + (1-\beta) \sum_{i=1}^n\p_{i}^{\delta}), 
\end{equation}
where $\beta=0.5$ is a weight, $\p_{0}^{\delta} = \|\po(k) - \po(k+1)\|$ and $\p_{i}^{\delta} = \|\p_i(k) - \p_i(k+1)\|$ are the travel distances of the payload position and each UAV, respectively. 
The sampler will converge to the minimum cost solution over time. 
The accepted samples by the collision checker ensure that the current formation distributes the load quasi-statically over the cables.

\subsubsection{Sampling Strategy}
Sampling $(\az_i,\el_i)$ uniformly and i.i.d. creates two major challenges: i) the probability of sampling a configuration that is collision-free without tangling of the cables decreases exponentially with $n$, and ii) the number of cable permutations that result in the same relative formation grows factorial in $n$. 
To mitigate this curse of dimensionality we propose a custom sampler, see \cref{alg:sampling}.
The sampler uses a preprocessing step to compute a set of witness cable configurations (\emph{formations}) that can be reached from the given valid start state $\x_s$.
During the actual sampling-based search, we rely on these witness formations.

For the preprocessing we initialize the witness set $\sS$ with the initial state $\x_s$ (\cref{alg:sampling:offline:init}).
For $M-1$ subsequent witnesses, we randomly choose a base state from the set (\cref{alg:sampling:offline:xb}) and uniformly sample a formation (\cref{alg:sampling:offline:xr}).
Then we solve an optimal assignment problem that minimizes the sum of the distances that the UAVs would have to move to change the formation from $\x^b$ to $\x^r$ (\cref{alg:sampling:offline:cost}).
For the cost, we extract the position of the UAVs as
\begin{equation}
    \operatorname{Pos}(\az_i, \el_i) = \po - l_i \qi(\az_i, \el_i),
\end{equation}
where $\qi(\az_i, \el_i)$ is given in \cref{eq:azeltoqi} and $\po$ is the position of the payload at the initial state $\x_s$ which is known to be collision-free.
The assignment problem can be solved optimally in polynomial time, for example by using the Hungarian Method (\cref{alg:sampling:offline:asg}).
We re-arrange the cables (\cref{alg:sampling:offline:shuffle}) and add the new witness if the whole formation change motion from $\x^b$ is collision-free (\cref{alg:sampling:offline:add}).

In the online search phase, we randomly pick a witness formation from the pre-computed set, add random noise to it, and augment it with a payload position uniformly drawn from the workspace $\W \subset \mathbb{R}^3$ (\crefrange{alg:sampling:online:start}{alg:sampling:online:end}).

\begin{algorithm}[t]
    \caption{Sampling Strategy}
    \label{alg:sampling}
    \DontPrintSemicolon

    \SetKwFunction{SampleUniform}{SampleUniform}
    \SetKwFunction{SampleGaussian}{SampleGaussian}
    \SetKwFunction{ChooseUniform}{ChooseUniform}
    \SetKwFunction{Pos}{Pos}
    \SetKwFunction{OptimalAssignment}{OptimalAssignment}

    \SetKwFunction{FOffline}{Preprocessing}
    \SetKwProg{Fn}{Function}{:}{}
    \Fn{\FOffline{$M$, $\x_s$, $\sigma$}}{
        $\sS = \{ \x_s = (\po, \az_1, \el_1, \hdots, \az_n, \el_n)^T \}$ \label{alg:sampling:offline:init}\;
        \While{$|\sS| < M$}{
            $\x^b \leftarrow \ChooseUniform(\sS)$ \label{alg:sampling:offline:xb} \Comment{pick a base state}
            $\x^r \leftarrow \SampleUniform(\C)$ \label{alg:sampling:offline:xr} \Comment{sample a new state}
            \tcc{Permute cables in $\x^r$ s.t. UAVs move minimally from $x^b$}
            $c[i,j] \leftarrow \|\Pos(\az_i^b, \el_i^b) - \Pos(\az_j^r, \el_j^r) \| \,\,\, \forall i,j \in \{1,\ldots,N\}$ \label{alg:sampling:offline:cost}\;
            $A \leftarrow \OptimalAssignment(c)$ \label{alg:sampling:offline:asg}\;
            $\x \leftarrow (\po, \az^r_{A[1]}, \el^r_{A[1]}, \hdots, \az^r_{A[n]}, \el^r_{A[n]})^T$ \label{alg:sampling:offline:shuffle}\;
            \If{$\x \in \C_{\mathrm{free}} \land (\x, \x^b) \in \C_{\mathrm{free}}$}{
                $\sS = \sS \cup \{ \x \}$ \label{alg:sampling:offline:add}
            }
        }
        \KwRet $\sS$
    }

    \SetKwFunction{FSampleState}{SampleState}
    \Fn{\FSampleState{$\sS$, $\sigma$}}{
        $\x \leftarrow \ChooseUniform(\sS)$ \label{alg:sampling:online:start}\;
        $(\_, \az_1, \el_1, \hdots, \az_n, \el_n) \leftarrow \SampleGaussian(\C, \x, \sigma)$
        \label{alg:sampling:online:gaussian}\;
        $\po \leftarrow \SampleUniform(\W)$ \label{alg:sampling:online:po}\;
        \KwRet $(\po, \az_1, \el_1, \hdots, \az_n, \el_n)$ \label{alg:sampling:online:end}\;
    }

\end{algorithm}
For the goal we only check if the payload reached the desired state. We use goal biasing and sample goal states in a similar fashion as in the online search (\crefrange{alg:sampling:online:start}{alg:sampling:online:gaussian}), except that the payload part is the user provided desired state rather than sampled as in \cref{alg:sampling:online:po}.

\begin{figure*}
    \centering
    \includegraphics[width=0.32\linewidth]{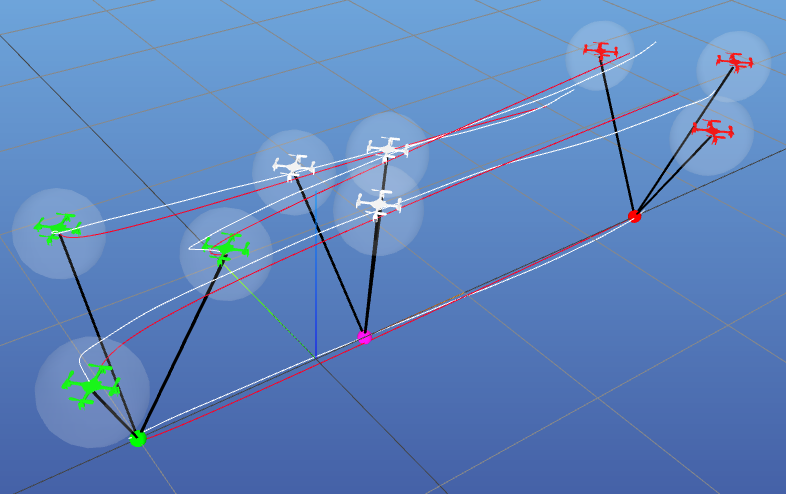}\hfill
    \includegraphics[width=0.32\linewidth]{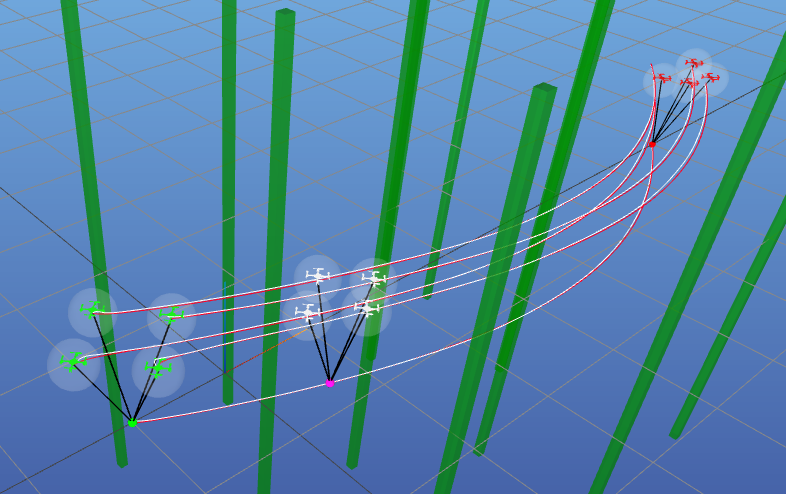}\hfill
    \includegraphics[width=0.32\linewidth]{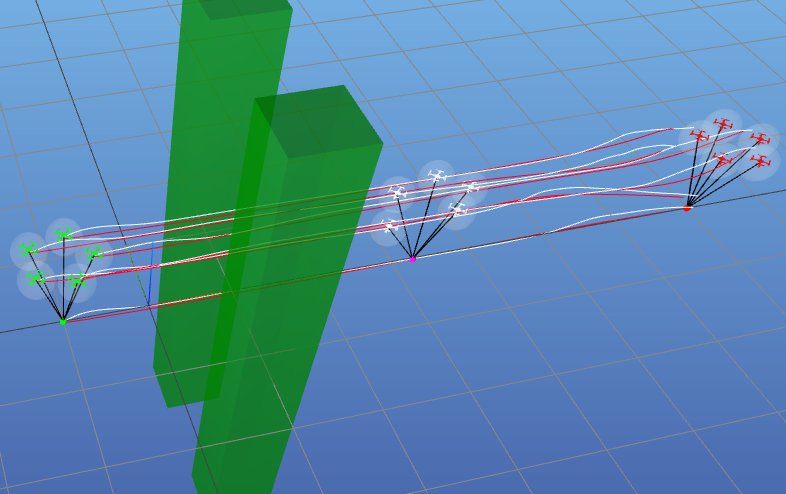}
    
    \caption{Validation scenarios. From left to right: \emph{empty} (3 robots), \emph{forest} (4 robots), \emph{window} (5 robots).
    Green UAVs (left on each picture) show the initial state, red UAVs the desired state. The red line represents the reference trajectory, and the white line is the tracked trajectory by our controller.
    For \emph{window}, the obstacles necessitate a formation change to pass through a narrow passage.
    }
    \label{fig:envs}
\end{figure*}

\subsubsection{Reference Trajectory (Offline)}
\label{sec:geom:ref:traj}
Since the geometric planner generates only $\xconfg$, the rest of the state vector $\x$ in \cref{eq:statespace} needs to be recovered. The payload velocity vector $\vo$ is computed with numerical differentiation over small time steps. The $\boldsymbol{\omega}_i$ is set to $\mathbf{0}$,  since we found that the numerical differentiation of $\qi$ is a noisy signal. Finally, the attitude states of the \emph{i}-th multirotor ($\R_i$, $\boldsymbol{\Omega}_i$) are set to $(\mathbf{I}_{3}, \mathbf{0})$, respectively, where $\mathbf{I}_{3}$ is the identity matrix.
\subsection{Nonlinear Trajectory Optimization (Offline)}
\label{sec:approach:opt}
The objective of the optimization step is to solve the full kinodynamic motion planning problem using the output of the geometric planner as initial guess.
While the geometric planner only plans for the payload and cable states, trajectory optimization considers the full state of the systems $\x$ and the motor forces $\uinp$ expressed in \cref{eq:statespace} and \cref{eq:motorforces}. 

To generate a valid initial guess, the geometric solution $\mathbf{X}_{\text{geom}}$ is interpolated with a small time discretization ($\Delta t = 0.01$ s), setting the velocity components to zero (which empirically results in a better initial guess) and using a default orientation for the robots. For the initial guess on the controls sequence, we use the constant motor forces that are required to hover each quadrotor individually.
The optimization problem is formulated as a nonlinear trajectory optimization
\begin{align}
\label{eq:opti}
    &\min_{\X, \U, \Delta t} \hspace{0.2cm} \sum_{k} ( \Delta t - \Delta t_0  )^2 + \beta_1 || \uinp_k ||^2  +  \nonumber \\
&  \qquad \qquad \qquad  \beta_2 || \ddot{\boldsymbol{x}}(\x_k, \uinp_k) ||^2, \\
    &\text{\noindent s.t.}\begin{cases}
    \x_{k+1} = \x_{k} \oplus  f(\x_k, \uinp_k) \Delta t \quad \forall k\in\{0, \ldots, T-1\},  \nonumber \\
     \uinp_k \in \mathcal{U},  \quad 
     \x_k  \in \mathcal{C} \quad \forall k\in\{0, \ldots, T\}, \\
     \x_0 = \x_s, \hspace{0.2cm} \x_T = \x_f, \nonumber \\
      g( \x_k ) \geq 0 \quad \forall k\in\{0, \ldots, T\}.
    \end{cases}
\end{align}
The decision variables are the sequence of controls $\mathbf{U} = \langle\uinp_0,\ldots, \uinp_{T-1} \rangle$ ,
the sequence of states $\mathbf{X} = \langle \x_0, ..., \x_{T} \rangle$ and the time-duration $\Delta t \in \mathbb{R}_+$ of the intervals in the time discretization. The number of steps $T$ is defined by the initial guess. The collision distance between the robots, payload, cables and environment is computed by  the signed distance function $g( \x_k )$. 

In the dynamics constraints, the continuous dynamics $f(\x_k, \uinp_k)$ are now discretized using Euler integration with a time interval subject to optimization. We use the notation $\oplus$ to highlight that 
some components of the state space $\x_k \in \mathcal{C}$  lie on manifolds (e.g. quaternions or unit vectors). 

The dynamics of the system is highly nonlinear, and the collision constraints define a non-convex feasible set, 
resulting in a very challenging nonlinear problem. To improve the robustness and success of trajectory optimization, 
and to ensure good local convergence towards a locally optimal trajectory, we combine three terms in the objective function.

The term  $( \Delta t - \Delta t_0  )^2 $ minimizes the time duration of the trajectory, with a small $\Delta t_0 $ acting as a proximal regularization. The term $|| \uinp_k ||^2$ 
minimizes the control effort.  The term $|| \ddot{\boldsymbol{x}}(\x_k,\uinp_k)||^2$ minimizes the acceleration of the system and is required to generate smooth trajectories and to ensure a good convergence of the solver. The coefficients $\beta_1,\beta_2>0$ are used to weigh the three terms. Together, these three terms combine the original objective of time minimization with a regularization that provides smooth gradients for the nonlinear optimizer to converge successfully.

The nonlinear trajectory optimization problem is solved using Differential Dynamic Programming (DDP) 
\cite{Crocoddyl, li2004iterative}, which ensures that the nonlinear dynamics constraints are always fulfilled. Goal constraints, states and control bounds are included with a squared penalty in the cost using the squared penalty method \cite{nocedal1999numerical}.

\subsection{QP-Force Allocation Geometric Controller (Online)} \label{sec:desCableForces}
As shown in \cref{fig:high-level-arch}, the QP-force allocation geometric controller tracks the generated motion plans.
Here, we demonstrate how to compute the required reference values of the controller from the motion planner output.

The formulation with QPs presented in \cite{wahba2023efficient} requires computing the reference cable forces $\muo$, that tracks the reference trajectory of the payload states and the reference cable formations $\qio$. 
Specifically, $\muo$ is derived from the reference trajectory and the reference motor forces $\uinp^i$ using $\muo = -T_i \qio$, where $T_i$ denotes the tension of the \(i\)-th cable.
The tension $T_i$ is calculated from the reference motor forces $\uinp^i$ as
\begin{equation}
\label{eq:cabletension}
    T_i = \| m_i(\ddot{\mathbf{p}}_{i_r} + g\ez) - f_{ci_{r}}\R_{i_r}\ez  \|,
\end{equation}
such that, $\ddot{\mathbf{p}}_{i_r}$ and $\R_{i_r}$ represent the reference acceleration and orientation of the \(i\)-th multirotor, and $f_{ci_{r}}$ is the collective thrust magnitude. Differentiating the position of the multirotor from \cref{eq:uavpos} twice yields the acceleration of the \(i\)-th multirotor as
\begin{equation}
    \label{eq:uavacc}
    \ddot{\mathbf{p}}_{i_r} = \poddr - l_i (\dotwio \times \qio + \wio \times \dotqio),
\end{equation}
and $\poddr$ represents the reference payload acceleration, and $\dotwio$ are determined using \cref{eq:dynamics}.
However, in the geometric case, we have $\boldsymbol{\omega}_i=\boldsymbol{\Omega}=\mathbf{0}$ and $\R_i=\mathbf{I}_{3}$, see \cref{sec:geom:ref:traj}.
Note that $\dotwio$ is not generally $\mathbf{0}$ and still computed using \cref{eq:dynamics}.
For the control input, we assume that each multirotor is hovering, i.e., $\boldsymbol{\eta}_i = (m_ig\mathbf{e}_3, 0,0,0)^T$.

After computing $\muo$, The cost function in the QPs can be modified as 
\begin{equation}
    \label{eq:newcost}
    c = \frac{1}{2}\normmu^2 + \lambda \| \muo - \mud\|^2,
\end{equation}
where the first term minimizes the sum of norms of the cable forces.
The second term minimizes the difference between the desired and the reference cable forces with $\lambda$ as a weighting factor.
Note that, even though in the geometric case the resulting $\muo$ may violate the dynamics, it is added as a soft constraint in \cref{eq:newcost}, thus the controller will still compute feasible $\mud$ to track. 

Finally, the low-level controller of each multirotor computes the motor forces that achieve these desired cable forces $\mud$, thus tracking the reference trajectory \cite{wahba2023efficient}.

\section{Experimental Results}

To validate the performance  of our method, we provide several real flight and simulation experiments.
In particular, we implement the sampling-based motion planner using OMPL~\cite{sucan2012open}, a widely used C++ library and rely on RRT* as sampling motion planning algorithm.
For optimization we extend Dynoplan~\cite{honig2022db}, an optimization-based motion planner based on Crocoddyl~\cite{Crocoddyl}, to include the system dynamics defined in \cref{eq:dynamics}.
Both geometric and optimization-based planners rely on FCL (Flexible Collision Library)~\cite{FCL} for collision checking.
We use sympy~\cite{sympy} to compute the analytical Jacobians of the dynamics and to generate efficient C code.

\subsection{Simulation Results}
For validation, we use a software-in-the-loop simulation where the flight controller code~\cite{wahba2023efficient} that runs on Bitcraze Crazyflie 2.1 multirotors is directly executed, together with the physics model that we implement in Dynobench~\cite{honig2022db}.
Due to the agile nature of multirotors, we use a small $\Delta t$ of \SI{0.01}{s} for both optimization and simulation.
All results are validated to fulfill the constraints in \cref{eq:general-optimization-problem}.

We test our motion planning approach on three different scenarios, see \cref{fig:envs} and the supplemental video: obstacle-free (i.e., Empty), a random forest-like environment, and a window environment, where the payload is transported through a narrow passage between two columns. The gap between each column and the origin linearly increases from $\SI{0.4}{m}$ to $\SI{0.5}{m}$ as the number of robots increases.
For each scenario, we evaluate five different problems increasing the number of robots, from two to six robots.

All scenarios for the motion planning experiments were solved on a workstation (AMD Ryzen Threadripper PRO 5975WX @ 3.6 GHz, 64 GB RAM, Ubuntu 22.04) and repeated 10 times. The runtime of the geometric motion planner was limited to \SI{300}{s}.

\subsubsection{Baseline Comparison}
\begin{table*}[ht]
    \caption{Simulation Results.
    Shown are mean values for the tracking error and energy usage of the final solution over 50 runs with a timelimit of \SI{300}{s} for the geometric planner. Standard deviation is small gray. Percentages are success rates.
    }
    \label{table:results}
\centering
\begin{tabular}{l|l||l|l|l|l|l|l}
Environment & Metrics &2 robots & 3 robots & 4 robots & 5 robots & 6 robots & success [\%]\\
\hline
\multirow{6}{*}{empty} & Error payload [m]  & {\bfseries 0.02} {\color{gray}\tiny 0.01 }  {\tiny {\bfseries 100} \% }  & 0.04 {\color{gray}\tiny 0.02 }  {\tiny {\bfseries 100} \% }  & 0.02 {\color{gray}\tiny 0.01 }  {\tiny {\bfseries 100} \% }  & {\bfseries 0.01} {\color{gray}\tiny 0.01 }  {\tiny 14 \% }  & \textemdash& 63 \\ & Error geom [m]  & 0.03 {\color{gray}\tiny 0.01 }  {\tiny {\bfseries 100} \% }  & 0.05 {\color{gray}\tiny 0.03 }  {\tiny {\bfseries 100} \% }  & 0.03 {\color{gray}\tiny 0.01 }  {\tiny 98 \% }  & 0.02 {\color{gray}\tiny 0.01 }  {\tiny 40 \% }  & \textemdash& 68 \\ & Error opt [m]  & {\bfseries 0.02} {\color{gray}\tiny 0.03 }  {\tiny 98 \% }  & {\bfseries 0.01} {\color{gray}\tiny 0.02 }  {\tiny {\bfseries 100} \% }  & {\bfseries 0.01} {\color{gray}\tiny 0.01 }  {\tiny {\bfseries 100} \% }  & {\bfseries 0.01} {\color{gray}\tiny 0.01 }  {\tiny {\bfseries 98} \% }  & {\bfseries 0.02} {\color{gray}\tiny 0.01 }  {\tiny {\bfseries 98} \% } & 99 \\\cline{2-7}
 & Energy payload [Wh]  & 0.05 {\color{gray}\tiny 0.00 }  & 0.08 {\color{gray}\tiny 0.00 }  & 0.10 {\color{gray}\tiny 0.00 }  & 0.12 {\color{gray}\tiny 0.00 }  & \textemdash\\ & Energy geom [Wh]  & 0.05 {\color{gray}\tiny 0.00 }  & 0.08 {\color{gray}\tiny 0.00 }  & 0.10 {\color{gray}\tiny 0.00 }  & 0.12 {\color{gray}\tiny 0.00 }  & \textemdash\\ & Energy opt [Wh]  & {\bfseries 0.04} {\color{gray}\tiny 0.00 }  & {\bfseries 0.05} {\color{gray}\tiny 0.00 }  & {\bfseries 0.07} {\color{gray}\tiny 0.00 }  & {\bfseries 0.11} {\color{gray}\tiny 0.01 }  & {\bfseries 0.13} {\color{gray}\tiny 0.01 } \\\hline
\multirow{6}{*}{forest} & Error payload [m]  & \textemdash & \textemdash & \textemdash & \textemdash & \textemdash& 0 \\ & Error geom [m]  & 0.08 {\color{gray}\tiny 0.06 }  {\tiny 46 \% }  & 0.10 {\color{gray}\tiny 0.06 }  {\tiny 66 \% }  & 0.10 {\color{gray}\tiny 0.06 }  {\tiny 62 \% }  & 0.11 {\color{gray}\tiny 0.06 }  {\tiny 16 \% }  & 0.20 {\color{gray}\tiny 0.08 }  {\tiny 2 \% } & 38 \\ & Error opt [m]  & {\bfseries 0.02} {\color{gray}\tiny 0.04 }  {\tiny {\bfseries 100} \% }  & {\bfseries 0.01} {\color{gray}\tiny 0.01 }  {\tiny {\bfseries 100} \% }  & {\bfseries 0.01} {\color{gray}\tiny 0.00 }  {\tiny {\bfseries 96} \% }  & {\bfseries 0.03} {\color{gray}\tiny 0.02 }  {\tiny {\bfseries 88} \% }  & {\bfseries 0.03} {\color{gray}\tiny 0.02 }  {\tiny {\bfseries 94} \% } & 96 \\\cline{2-7}
 & Energy payload [Wh]  & \textemdash & \textemdash & \textemdash & \textemdash & \textemdash\\ & Energy geom [Wh]  & {\bfseries 0.05} {\color{gray}\tiny 0.00 }  & 0.08 {\color{gray}\tiny 0.00 }  & 0.10 {\color{gray}\tiny 0.00 }  & {\bfseries 0.12} {\color{gray}\tiny 0.00 }  & 0.15 {\color{gray}\tiny 0.00 } \\ & Energy opt [Wh]  & {\bfseries 0.05} {\color{gray}\tiny 0.00 }  & {\bfseries 0.07} {\color{gray}\tiny 0.00 }  & {\bfseries 0.09} {\color{gray}\tiny 0.00 }  & {\bfseries 0.12} {\color{gray}\tiny 0.01 }  & {\bfseries 0.14} {\color{gray}\tiny 0.01 } \\\hline
\multirow{6}{*}{window} & Error payload [m]  & 0.04 {\color{gray}\tiny 0.02 }  {\tiny 4 \% }  & 0.06 {\color{gray}\tiny 0.04 }  {\tiny 12 \% }  & 0.04 {\color{gray}\tiny 0.02 }  {\tiny 2 \% }  & \textemdash & \textemdash& 4 \\ & Error geom [m]  & 0.10 {\color{gray}\tiny 0.09 }  {\tiny 34 \% }  & 0.10 {\color{gray}\tiny 0.06 }  {\tiny 14 \% }  & \textemdash & \textemdash & \textemdash& 10 \\ & Error opt [m]  & {\bfseries 0.02} {\color{gray}\tiny 0.02 }  {\tiny {\bfseries 98} \% }  & {\bfseries 0.01} {\color{gray}\tiny 0.01 }  {\tiny {\bfseries 100} \% }  & {\bfseries 0.02} {\color{gray}\tiny 0.01 }  {\tiny {\bfseries 90} \% }  & {\bfseries 0.04} {\color{gray}\tiny 0.03 }  {\tiny {\bfseries 54} \% }  & {\bfseries 0.05} {\color{gray}\tiny 0.03 }  {\tiny {\bfseries 56} \% } & 80 \\\cline{2-7}
 & Energy payload [Wh]  & {\bfseries 0.05} {\color{gray}\tiny 0.00 }  & 0.08 {\color{gray}\tiny 0.00 }  & 0.10 {\color{gray}\tiny 0.00 }  & \textemdash & \textemdash\\ & Energy geom [Wh]  & {\bfseries 0.05} {\color{gray}\tiny 0.00 }  & 0.08 {\color{gray}\tiny 0.00 }  & \textemdash & \textemdash & \textemdash\\ & Energy opt [Wh]  & {\bfseries 0.05} {\color{gray}\tiny 0.00 }  & {\bfseries 0.07} {\color{gray}\tiny 0.00 }  & {\bfseries 0.09} {\color{gray}\tiny 0.01 }  & {\bfseries 0.13} {\color{gray}\tiny 0.01 }  & {\bfseries 0.15} {\color{gray}\tiny 0.01 } \\
\end{tabular}
\end{table*}

We consider three different approaches for payload transport:
\emph{\textbf{Payload}} uses geometric planning for the payload alone and the resulting trajectory is tracked using the specialized payload transport controller.
\emph{\textbf{Geom}} uses geometric planning for the payload, cables, and UAVs as described in \cref{sec:approach:geometric}. The same controller can then use the augmented cost function \cref{eq:newcost} to track both the cables and the payload, allowing formation changes to pass through passages.
 \emph{\textbf{Opt}} uses the proposed pipeline for kinodynamic motion planning: We first generate a geometric plan for the cable-payload system,
 (\cref{sec:approach:geometric})
 and then we use the nonlinear trajectory optimization to generate feasible trajectories
 (\cref{sec:approach:opt}). 

We analyze tracking error, energy usage, and success rate in 15 different settings. The results are shown in  \cref{table:results}.
We consider an execution a success if i) a reference trajectory is successfully computed, ii) no collisions occurred when the controller is tracking this reference trajectory, and iii) the controller reaches the goal.
For the empty environment, all algorithms are successful. In the forest environment, the simple \emph{payload} approach results in collisions since the UAVs are ignored during planning, and the planned path tends to be too close to obstacles.
Here, the \emph{geom} approach works in the majority of the tested cases.
For the \emph{window} case, only our kinodynamic planner is highly successful, since the geometric planner often produces desired cable states that cannot be tracked accurately when considering the kinodynamic constraints.
\begin{figure}
    \centering
    \includegraphics[height=2.9cm]{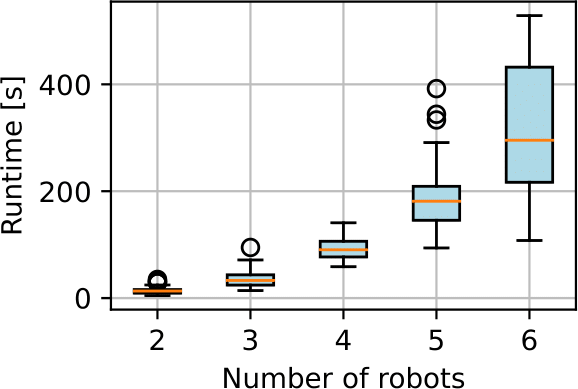}\hfill
    \includegraphics[height=3.0cm]{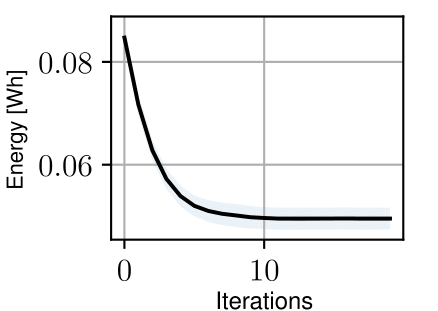}
    \caption{Left: Computational effort in seconds for the optimization to compute a solution in the forest environment over different numbers of robots. Right: Solution quality (in terms of energy) when sequentially solving the kinodynamic optimization over multiple iterations.}
    \label{fig:cost_opt_and_iter_opt}
\end{figure}
\begin{figure}[t]
    \centering
    \includegraphics[width=0.8\linewidth]{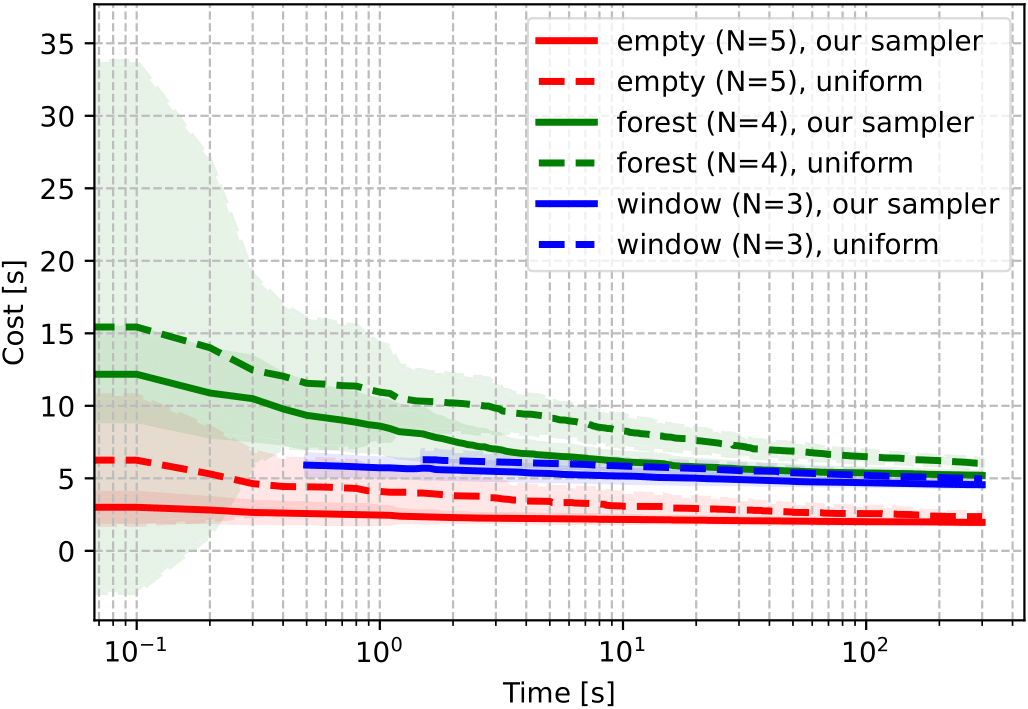}
    \caption{Examples for the sampling-based geometric planner using different environment and sampling strategies. The plot shows the mean and standard deviation (shaded) for cost convergence over runtime (log-scale), if the success rate is over $50 \%$ (50 trials).}
    \label{fig:cost_geom}
\end{figure}
The average tracking error in all settings is significantly lower for our approach \emph{opt} (up to 9 times lower compared to the geometric solution), since the full system states are considered.
We also compute the expected energy consumption of the flight using a model for the Bitcraze Crayflie 2.X multirotors \footnote{\url{https://bitcraze.io/documentation/repository/crazyflie-firmware/master/functional-areas/pwm-to-thrust/}}.
Here, the energy linearly depends on the flight time as well as the forces $\uinp^i$ that the propellers create.
Not surprisingly, the energy for \emph{payload} and \emph{geom} are almost identical, while our method, \emph{opt}, reaches a reduction of around 10 \%.
However, the energy is reduced by almost 50 \% with iterative optimization (see \cref{sec:iter}).
Moreover, the energy usage almost linearly increases with the number of robots, which is not surprising as multirotors require a lot of energy just to stay airborne. 

\begin{figure*}[ht]
  \centering
    \includegraphics[width=0.4\textwidth]{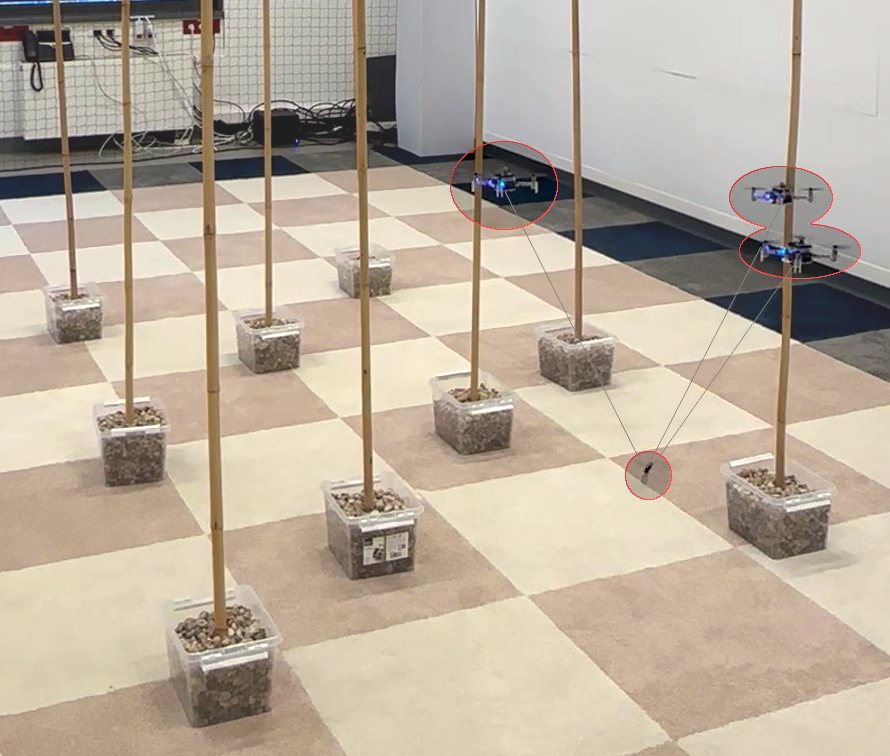}
    \includegraphics[width=0.4\textwidth]{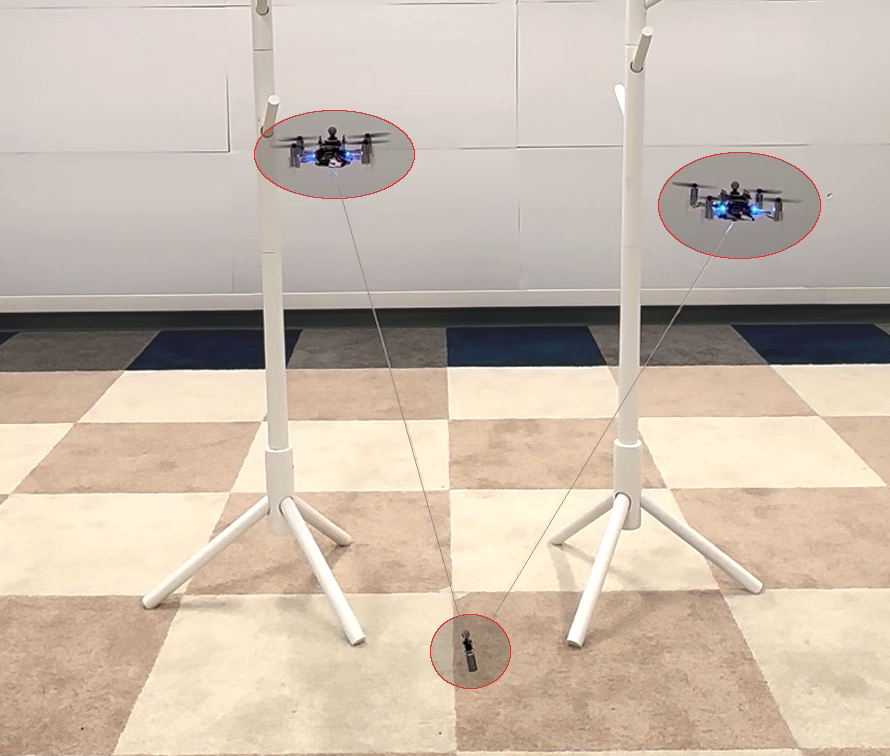}
  \caption{Real flights validation scenarios. left: forest (3 robots), right: window (2 robots). The payload in both scenarios is modeled as a \SI{10}{g} point mass. }
  \label{fig:real_flights}
\end{figure*}

\begin{table}[ht]
\centering
    \caption{Real Flights Results.
    Shown are mean values for the tracking error, energy usage and the planning time of each approach of 10 flight experiments executed on-board of the CFs to track the reference trajectories. Standard deviation is small gray. Percentages are success rates.
    }
\label{table:real_flights}
\begin{tabular}{l|l|l|l}
Env. & Metric &2 robots & 3 robots\\
\hline
\multirow{6}{*}{window} & Error geom [m] &  { 0.062}   {\color{gray}\tiny 0.04 }  {\tiny 80 \% }  &  { 0.121}   {\color{gray}\tiny 0.06 }  {\tiny 60 \% }\\ & Error opt [m] &  { \textbf{0.056}}   {\color{gray}\tiny 0.03 }  {\tiny 100 \% }  &  { \textbf{0.085}}   {\color{gray}\tiny 0.04 }  {\tiny 100 \% }\\\cline{2-4}
 & Energy geom [Wh]  &  { 0.022}   {\color{gray}\tiny 0.00 }  &  { 0.027}   {\color{gray}\tiny 0.00 } \\ & Energy opt [Wh]  &  { \textbf{0.016}}   {\color{gray}\tiny 0.00 }  &  { \textbf{0.020}}   {\color{gray}\tiny 0.00 } \\\cline{2-4}
& Plan time geom [s] &   { 0.4}   {\color{gray}\tiny 0.42}  &  { 0.1}   {\color{gray}\tiny  0.05}  \\
& Plan time opt [s] &  { 26.8}   {\color{gray}\tiny 9.45 }   &  { 46.0}   {\color{gray}\tiny 12.5 } \\\hline
\multirow{6}{*}{forest} & Error geom [m] &  { 0.092}   {\color{gray}\tiny 0.08 }  {\tiny 60 \% }  &  \hspace{0.1cm}--- \\ & Error opt [m] &  { \textbf{0.057}}   {\color{gray}\tiny 0.04 }  {\tiny 100 \% }  &  { \textbf{0.118}}   {\color{gray}\tiny 0.05 }  {\tiny 100 \% } \\\cline{2-4}
 & Energy geom [Wh]  &  { 0.017}   {\color{gray}\tiny 0.00 }  &  \hspace{0.1cm}--- \\ & Energy opt [Wh]  &  { \textbf{0.012}}   {\color{gray}\tiny 0.00 }  &  { \textbf{0.024}}   {\color{gray}\tiny 0.01 } \\\cline{2-4}
 & Plan. time geom [s] &  { 0.3}   {\color{gray}\tiny 0.43}  &  { 168.4}   {\color{gray}\tiny 105.6} \\
& Plan. time opt [s] &  { 22.2}   {\color{gray}\tiny 3.71} &  { 57.8}   {\color{gray}\tiny 16.13 } \\
\end{tabular}
\end{table}

\subsubsection{Computational Effort}
We analyze the runtime of our two planning phases, geometric planning and optimization, separately. For the geometric planner, we show that our new sampling strategy (\cref{alg:sampling}) compared to the uniform sampling, significantly reduces the time until a low-cost solution is reached, see \cref{fig:cost_geom}. These results are more significant for the environments with more open space (\emph{empty} and \emph{window}). Moreover, our results show that the sampling strategy effectively reduces the standard deviation of the cost, making it more likely that we find a high-quality solution quickly.

For the optimization, we are interested in the scalability with the number of robots. The number of decision variables is linear in the time horizon and the number of robots. 
Theoretically, trajectory optimization using DDP scales cubically with the state dimensionality (i.e., the number of robots) and linearly with the time horizon. However, we observe that adding robots also results in more nonlinear iterations, increasing the overall computation time, see \cref{fig:cost_opt_and_iter_opt}.

\subsubsection{Iterative Optimization (Offline)}
\label{sec:iter}
To minimize the trajectory duration, we solve a sequence of optimization problems for decreasing values of $\Delta t_0$ \eqref{eq:opti} using the solutions of each problem as initial guess for the next one.
Examples of this approach are apparent in sequential convex programming (SCP)~\cite{malyutaConvexOptimizationTrajectory2022a} as well as prior motion planning techniques for multirotors~\cite{wolfgang2018}.
We observe the benefits in our application, as shown in \cref{fig:cost_opt_and_iter_opt} for the \emph{forest} environment with 4 robots, the estimated energy consumption of the team is reduced by almost 50 \% just by repeating the optimization ten times.

\subsection{Real Flights Results}
\subsubsection{Physical Setup} 
To provide concrete validation to our simulation results, we conduct several real flight tests. For the real platform, we use multirotors of type Bitcraze Crazyflie 2.1 (CF), where we use the same flight controller code \cite{wahba2023efficient} running on-board as in SITL. These are small (9 cm rotor-to-rotor) and lightweight (\SI{34}{g}) products that are commercially available.
Controller and extended Kalman filter for state estimation run directly on-board the STM32-based flight controller (168 MHz, 192 kB RAM). For all scenarios, we use dental floss as cables with length \SI{0.5}{m} and the payload mass is \SI{10}{g}. We use magnets to connect the cables and the payload/multirotors to be easily repaired. On the host side, we use Crazyswarm2, which is based on Crazyswarm~\cite{preiss2017crazyswarm} but uses ROS~2~\cite{macenski2022robot} to control and send commands for multiple CFs.
In particular, we equip each multirotor and the payload with a single reflective marker for position tracking at $\SI{100}{Hz}$ using an OptiTrack motion capture system in a $7.5\times4\times2.75$ \si{m^3} flight space. 
\subsubsection{Results}
We validate the functionality and the quality of our kinodynamic planner \emph{opt} against \emph{geom} through especially designed environments, where the state-of-the-art motion planners fail to generate feasible trajectories for the full system that can be tracked by the controller \cite{wahba2023efficient}. 
As shown in \cref{fig:real_flights}, we test the experiments in the window and forest environments for two and three robots. Similar to our prior work \cite{wahba2023efficient}, we only fly up to three robots 
due to the computationally constrained microcontroller.

We can generate solutions within a few minutes, see \cref{table:real_flights}, where in most cases the optimization is the slower part. In the forest case with 3 robots, the geometric planning is computationally expensive due to a high obstacle density. Note that both window and forest examples are different from the simulation results to match our flight space constraints.

As in the simulation evaluation, we compare the tracking error and the energy usage for ten executed flights. Moreover, we compare the success rate of each method in tracking the reference trajectory to the goal state $\x_g$ while avoiding inter-robot or robot/obstacle collisions. As shown in \cref{table:real_flights}, \emph{opt} is succeeding in all cases.
However, when tracking the plans generated by the geometric planner we observed failures either by the team crashing into obstacles or the whole team becoming unstable due to infeasibility of the provided reference.
In the 3 robot forest case, such tracking failures could even be observed in simulation, explaining the \SI{0}{\percent} success rate.

\section{Conclusion}
We propose a hierarchical kinodynamic motion planning algorithm for the cable-suspended payload transport system.
Our method directly considers obstacles, inter-robot collisions, the full dynamics, the actuation limits and allows us to plan feasible reference trajectories that can be tracked accurately by our controller in realtime.
We compare our method with the state-of-the-art baselines in multiple simulation and real experiments. 
In all cases, we achieve higher success rates and enhance the energy consumption of the executed motions, thereby maximizing effectiveness.

In the future, we would like to extend our method to the rigid payload and enable realtime planning with an obstacle-aware controller.
\balance
\printbibliography

\end{document}